# A NOVEL MULTICRITERIA GROUP DECISION MAKING APPROACH WITH INTUITIONISTIC FUZZY SIR METHOD

## JUNYI CHAI and JAMES N.K. LIU

*Department of Computing,*
*The Hong Kong Polytechnic University*
*Hung Hom, Kowloon, Hong Kong, SAR*
*{ csjchai, csnkliu }@comp.polyu.edu.hk*

ABSTRACT - The superiority and inferiority ranking (SIR) method is a generation of the well-known PROMETHEE method, which can be more efficient to deal with multi-criterion decision making (MCDM) problem. Intuitionistic fuzzy sets (IFSs), as an important extension of fuzzy sets (IFs), include both membership functions and non-membership functions and can be used to, more precisely describe uncertain information. In real world, decision situations are usually under uncertain environment and involve multiple individuals who have their own points of view on handing of decision problems. In order to solve uncertainty group MCDM problem, we propose a novel intuitionistic fuzzy SIR method in this paper. This approach uses intuitionistic fuzzy aggregation operators and SIR ranking methods to handle uncertain information; integrate individual opinions into group opinions; make decisions on multiple-criterion; and finally structure a specific decision map. The proposed approach is illustrated in a simulation of group decision making problem related to supply chain management.

Keyword: Intuitionistic fuzzy set; SIR method; Group MCDM; Supply chain management

## 1. INTRODUCTION

Multi-Criterion Decision Making (MCDM) is a process in which decision makers evaluate each alternative according to multiple criteria. Many representative methods are introduced to solve MCDM problem in business and industry areas, including AHP, TOPSIS, UTA, PROMETHEE, ELECTRE, etc. These methods can be generally divided into three categories: (1) multi-criteria utility theory, (2) outranking relations and (3) preference disaggregation. Particularly, Xu [1] extended PROMETHEE with a Superiority and Inferiority Ranking (SIR) method which integrated the outranking approach and the concept of TOPSIS method. However, a drawback of these approaches is that they mostly just consider the decision making with certain information of the weights and decision values, which makes them much less useful when managing uncertain or fuzzy knowledge.

Intuitionistic Fuzzy Sets (IFSs) [2] extend the concept of the membership functions in Fuzzy Sets (FSs) [3] to include the non-membership functions and degrees of hesitation. Further theoretical work has been provided by Chen and Tan [4] who defined the score function and Hong and Choi [5] who defined the accuracy function in vague sets [6], and Bustince and Burillo [7] proved that vague sets are IFSs. Szmidt and Kacprzyk [8] proposed a nonprobabilistic-type entropy measure; Deschrijver et al. [9] introduced the concepts of $t$-norm and $t$-conorm. More recently, Xu [10] developed some IFs operators (e.g. IFWA, IFWG, IFHA, IFHG) for aggregating intuitionistic fuzzy information. The application of IFSs has been seen in design schemes selection [11], web services selection [12], and supplier selection problems [13].

This paper proposes a novel multi-criterion group decision making approach to address the supply chain partner selection problem under uncertain environment. We provide an Intuitionistic Fuzzy Superiority and Inferiority Ranking (IF-SIR) method for application in group decision-making. In facing the uncertain situations that the individual decision value and the weights of criteria and decision makers are all just provided by fuzzy natural language terms, we first define these terms through intuitionistic fuzzy sets. We then use IFWA operator and IFWG operator to integrate individual opinions into group opinions. After that, we set a threshold function to obtain the Superiority/Inferiority flow of every alternative. Finally, we order alternatives according to IF-SIR ranking rules and map them into a final decision. We illustrate this procedure with an example and simulate it using the MATLAB tool. The experiment indicates that this proposed method can solve the uncertainty group MCDM problems.

The rest of this paper is organized as follow. Section 2 presents some basic definitions in intuitionistic fuzzy sets. Section 3 proposes a detailed description of the intuitionistic fuzzy SIR method. Section 4 provides an illustration of the MCDM problem with reference to the area of supply chain management. Section 5 gives our conclusion and outlines directions for future work.

## 2. INTUITIONISTIC FUZZY SETS: DEFINITIONS

Some basic definitions of Intuitionistic Fuzzy Sets (IFSs) are presented in this section:

(1) Intuitionistic fuzzy set $A$ in a finite set $X$ ($X = \emptyset$) can be written as
$A = \{<x, \mu_A(x), v_A(x)> | x \in X\}$, where: $0 \leq \mu_A + v_A \leq 1$, $x \in X$;
$\mu_A : X \to [0,1], x \in X \to \mu_A(x) \in [0,1]$; $v_A : X \to [0,1], x \in X \to v_A(x) \in [0,1]$;
The hesitation degrees: $\pi_A = 1 - \mu_A - v_A$, $x \in X$.

(2) Let $a_1$ and $a_2$ be Intuitionistic Fuzzy Number (IFN) of the set $X$. Other basic operations of IFSs [2] include:
Complement: $\bar{a} = (v_a, \mu_a)$      Eq. (1)
Addition: $a_1 \oplus a_2 = \{\mu_{a_1} + \mu_{a_2} - \mu_{a_1}\mu_{a_2}, v_{a_1}v_{a_2}\}$; Multiplication: $a_1 \otimes a_2 = \{\mu_{a_1}\mu_{a_2}, v_{a_1} + v_{a_2} - v_{a_1}v_{a_2}\}$

(3) De [14] defined two further operations:
Multiple law: $\lambda a = (1-(1-\mu_a)^\lambda, v_a^\lambda)$, $\lambda > 0$; Exponent law: $a^\lambda = (\mu_a^\lambda, 1-(1-v_a)^\lambda)$, $\lambda > 0$

(4) Chen and Tan [4] defined the Score Function: $s(a) = \mu_a - v_a$.

(5) Hong and Choi [5] defined the Accuracy Function: $h(a) = \mu_a + v_a$.

(6) Xu [10] described the IFWA operator and IFWG operator, and developed a method for comparing two intuitionistic fuzzy numbers using $s(a)$ and $h(a)$:
If $s(a_1) < s(a_2)$, then, $a_1 < a_2$      Eq. (2)
If $s(a_1) = s(a_2)$, then, 1) If $h(a_1) = h(a_2)$, then $a_1 = a_2$  2) If $h(a_1) < h(a_2)$, then $a_1 < a_2$
                           3) If $h(a_1) > h(a_2)$, then $a_1 > a_2$

The group MCDM problem involves multiple individuals assessing alternatives based on multiple-criteria. In this paper, decision factors in group MCDM problem are formulated as shown in Table I.

| Decision factors | Formulation |
|---|---|
| Alternative sets | $Y_i = \{Y_1, Y_2, ..., Y_n\}$   $i = 1, 2, ..., n$ |
| Decision maker sets | $e_k = \{e_1, e_2, ..., e_l\}$   $k = 1, 2, ..., l$ |
| Criterion sets | $G_j = \{G_1, G_2, ..., G_m\}$   $j = 1, 2, ..., m$ |
| Decision maker weights | $w_k = (w_1, w_2, ..., w_l)^T$   $w_k = (\mu_k, v_k, \pi_k)$ |
| Criterion weights | $\omega_j = (\omega_1, \omega_2, ..., \omega_m)^T$   $\omega_j^{(k)} = (\mu_j^{(k)}, v_j^{(k)})$, |
| Individual decision matrix | $d_{ij}^{(k)} = (\mu_{ij}^{(k)}, v_{ij}^{(k)})$, |
| Group-integrated decision information | Matrix: $\bar{d}_{ij} = (\bar{\mu}_{ij}, \bar{v}_{ij})$, Weight: $\bar{\omega}_j = (\bar{\mu}_j, \bar{v}_j)$, , |

**Table I. Formulation of decision factors.**

## 3. INTUITIONISTIC FUZZY SIR METHOD

The intuitionistic fuzzy SIR method is given as following steps:

**Step 1.** Determine the individual measure degree $\xi_k$

The weights of decision makers are assigned in fuzzy natural language terms which are defined using intuitionistic fuzzy sets. Table II provides an example of the term measured on "Importance" and "Quality" at different levels.

| Level | "Importance" Measure | "Quality" Measure | IFNs |
|---|---|---|---|
| L1 | Extremely Important(EI) | Extremely Positive(EP) | (1.00, 0.00) |
| L2 | Great Important(GI) | Absolutely Positive(AP) | (0.90, 0.10) |
| L3 | Very Important(VI) | Very Very Positive(VVP) | (0.80, 0.10) |
| L4 | Important(I) | Very Positive(VP) | (0.70, 0.20) |
| L5 | Medium(M) | Positive(P) | (0.60, 0.30) |
| L6 | Less Important(LI) | Medium(M) | (0.50, 0.40) |
| L7 | Unimportant(U) | Negative(N) | (0.40, 0.50) |
| L8 | Not Important(NI) | Very Negative(VN) | (0.05, 0.80) |
| L9 | Unconsidered(UC) | Extremely Negative(EN) | (0.00, 0.10) |

**Table II. "Importance" and "Quality" ranked in Intuitionistic Fuzzy Sets**

Let $w_k = (\mu_k, v_k, \pi_k)$ be the IFNs confirmed based on Table II.
The Normalized Euclidean Distance [15] is obtained by:

$$D_k(w_k, w^+) = \sqrt{\frac{1}{2}\left((\mu_k - \mu^+)^2 + (v_k - v^+)^2 + (\pi_k - \pi^+)^2\right)} \text{ where } w^+ = (\mu^+, v^+, \pi^+) = (1, 0, 0). \quad \text{Eq. (3)}$$

The Similarity Measure [16] is then obtained by:

$$\xi_k(w_k, w^+) = D_k(w_k, w^+)/D_k(w_k, \overline{w}^+) \text{ where } 0 \leq \xi_k(w_k, w^+) \leq +\infty, \text{ and } \overline{w}^+ = w^-$$

In order to form this function into interval [0, 1], we transfer it as following function:

$$\xi_k(w_k, w^+) = 1 - \frac{D_k(w_k, w^+)}{D_k(w_k, w^+) + D_k(w_k, w^-)} = \frac{D_k(w_k, w^-)}{D_k(w_k, w^+) + D_k(w_k, w^-)} \quad \text{Eq. (4)}$$

where $0 \leq \xi_k(w_k, w^+) \leq 1$ and if $\xi_k(w_k, w^+) \to 1$, then $w_l \to w^+ = (1, 0, 0)$.

Finally, we obtain the vector of real numbers, $\xi_k = (\xi_1, \xi_2, ..., \xi_l)^T$ as individual measure degrees.

**Step 2.** Group integration

We use IFWA operator and IFWG operator to integrate individual opinions into group opinions as following.

a) Individual decision matrix integration.

$$\overline{d}_{ij} = IFWA_{\xi_k}(d_{ij}^{(1)}, d_{ij}^{(2)}, ..., d_{ij}^{(l)}) = \xi_1 d_{ij}^{(1)} \oplus \xi_2 d_{ij}^{(2)} \oplus ... \oplus \xi_l d_{ij}^{(l)}$$

$$= (1 - \prod_{k=1}^{j}(1 - \mu_{ij}^{(k)})^{\xi_k}, \prod_{k=1}^{j}(v_{ij}^{(k)})^{\xi_k}) = (\overline{\mu}_{ij}, \overline{v}_{ij}) \quad \text{Eq. (5)}$$

b) Individual criterion weights integration.

$$\overline{\omega}_j = IFWG_{\xi_k}(\omega_j^{(1)}, \omega_j^{(2)}, ..., \omega_j^{(k)}) = (\omega_j^{(1)})^{\xi_1} \otimes (\omega_j^{(2)})^{\xi_2} \otimes ... \otimes (\omega_j^{(l)})^{\xi_l}$$

$$= (\prod_{k=1}^{l}(\mu_j^{(k)})^{\xi_k}, 1 - \prod_{k=1}^{l}(1 - v_j^{(k)})^{\xi_k}) = (\overline{\mu}_j, \overline{v}_j) \quad \text{Eq. (6)}$$

From this step, we obtain group-integrated decision matrix $\overline{d}_{ij} = (\overline{\mu}_{ij}, \overline{v}_{ij})$ and the criterion weights $\overline{\omega}_j = (\overline{\mu}_j, \overline{v}_j)$.

**Step 3.** Obtain intuitionistic fuzzy Superiority/Inferiority matrix

**a)** Confirm the performance function $g(Y_i)$

The group-integrated decision matrix can be written as $d_{ij} = (\mu_{ij}, v_{ij}, \pi_{ij}) = (\overline{\mu}_{ij}, \overline{v}_{ij}, 1 - \overline{\mu}_{ij} - \overline{v}_{ij})$.

We define $g(Y_i)$ as the performance function. Using the Normalized Hamming Distance and Similarity Measure [15, 16], we can calculate $g(Y_i)$ by:

$$D_j(d_{ij}, d^+) = \frac{1}{2}\left(|\mu_{ij} - \mu^+| + |v_{ij} - v^+| + |\pi_{ij} - \pi^+|\right) \quad \text{Eq. (7)}$$

$$g(Y_i) = \psi(d_{ij}, d^+) = \frac{D_j(d_{ij}, d^-)}{D_j(d_{ij}, d^+) + D_j(d_{ij}, d^-)}, \text{ Here, } 0 \leq g(Y_i) \leq 1, \text{ and if } g(Y_i) \to 1, \text{ then } d_{ik} \to d^+. \quad \text{Eq. (8)}$$

**b)** Confirm the preference intensity $P_k(Y_i, Y_t)$

We define $P_k(Y_i, Y_t)$ as the preference intensity of alternative $Y_i$ over alternative $Y_t$, with respect to the $k$th criterion.

$$P_k(Y_i, Y_t) = \phi_k(g_k(Y_i) - g_k(Y_t)) = \phi_k(d) \quad i, t = 1, 2, ..., n, i \neq t, k = 1, 2, ..., l \quad \text{Eq. (9)}$$

where $\phi_k(d)$ is a non-decreasing function from the real number R to [0, 1]. Generally, $\phi_k(d)$ can be chosen from six generalized threshold functions [17], or defined by decision makers themselves.

**c)** Obtain Superiority matrix and Inferiority matrix

For alternative $Y_i$, we obtain the intuitionistic fuzzy Superiority/Inferiority index and matrices.

Superiority index (S-index): $S_k(Y_i) = \sum_{i=1}^{n} P_k(Y_i, Y_t) = \sum_{i=1}^{n} \phi_k(g(Y_i) - g(Y_t))$; S-matrix: $S = (S_k(Y_i))_{n \times m}$ Eq. (10)

Inferiority index (I-index): $I_k(Y_i) = \sum_{i=1}^{n} P_k(Y_t, Y_i) = \sum_{i=1}^{n} \phi_k(g(Y_t) - g(Y_i))$; I-matrix: $I = (I_k(Y_i))_{n \times m}$ Eq. (11)

**Step 4.** Obtain Superiority flow and Inferiority flow

We calculate the weighted Superiority flow and Inferiority flow as follows:

S-flow: $\phi^>(Y_i) = \sum_{j=1}^{m} \overline{\omega}_j S_j(Y_i) = IFWA_{S_j(Y_i)}(\overline{\omega}_1, \overline{\omega}_2, ..., \overline{\omega}_m)$

$$= (1 - \prod_{j=1}^{m}(1-\bar{\mu}_j)^{S_j(Y_i)}, \prod_{j=1}^{m} \bar{v}_j^{S_j(Y_i)}) = (\mu^>(Y_i), v^>(Y_i)) \qquad \text{Eq. (12)}$$

I-flow: $\phi^<(Y_i) = \sum_{j=1}^{n} \bar{\omega}_j I_j(Y_i) = IFWA_{I_j(Y_i)}(\bar{\omega}_1, \bar{\omega}_2, ..., \bar{\omega}_m)$

$$= (1 - \prod_{j=1}^{m}(1-\bar{\mu}_j)^{I_j(Y_i)}, \prod_{j=1}^{m} \bar{v}_j^{I_j(Y_i)}) = (\mu^<(Y_i), v^<(Y_i)) \qquad \text{Eq. (13)}$$

Therefore, we obtain S-flow and I-flow of alternative $Y_i$ as $Y_i(\phi^>(Y_i), \phi^<(Y_i))$.

Clearly, when the higher S-flow $\phi^>(Y_i)$ and the lower I-flow $\phi^<(Y_i)$, alternative $Y_i$ is better.

**Step 5.** The intuitionistic fuzzy SIR ranking rules

a) Confirm Superiority ranking and Inferiority ranking

The descending order of $\phi^>(Y_i)$ is used to obtain the Superiority ranking (called S-ranking) by rule:

$Y_i P_> Y_k$ iff $\phi^>(Y_i) > \phi^>(Y_k)$ and $Y_i I_> Y_k$ iff $\phi^>(Y_i) = \phi^>(Y_k)$

Similarly, the ascending order of $\phi^<(Y_i)$ is used to obtain the Inferiority ranking (called I-ranking) by rule:

$Y_i P_< Y_k$ iff $\phi^<(Y_i) > \phi^<(Y_k)$ and $Y_i I_< Y_k$ iff $\phi^<(Y_i) = \phi^<(Y_k)$

b) Confirm the IF-SIR ranking

Combine S-ranking and I-ranking into a partial ranking structure $R^* = \{P, I, R\} = R_>^* \cap R_<^*$ of two alternatives $Y_i$ and $Y_k$ by applying the intersection principles:

(1) The Preference relation $P$ by rule:
$Y_i P Y_k$ iff ($Y_i P_> Y_k$ and $Y_i P_< Y_k$) or ($Y_i P_> Y_k$ and $Y_i I_< Y_k$) or ($Y_i I_> Y_k$ and $Y_i P_< Y_k$)

(2) The Indifference relation $I$ by rule: $Y_i I Y_k$ iff ($Y_i I_> Y_k$ and $Y_i I_< Y_k$)

(3) The Incomparability relation $R$ by rule: $Y_i R Y_k$ iff ($Y_i P_> Y_k$ and $Y_k P_< Y_i$) or ($Y_k P_> Y_i$ and $Y_i P_< Y_k$)

**Step 6.** Map the complete ranking and make decision

After determined every partial ranking structure of alternatives, we can map the complete IF-SIR ranking (including the relationships between any two of alternatives) for decision making.

## 4. ILLUSTRATIVE EXAMPLE

The major aims of Supply Chain Management (SCM) are to reduce business risk and production costs and improve customer services. There have been many studies on the problem of choosing supply chain partners [18], which is essentially a Multi-criterion Decision Making (MCDM) problem in which decision makers evaluate each alternative according to multiple criteria. With this in mind, suppose a company wishes to choose supply chain partner and five alternative companies ($Y_i, i = 1,2,3,4,5$) are being considered according to four criteria ($G_j, j = 1,2,3,4$): Financial Situation ($G_1$); Technology Performance ($G_2$); Management Performance ($G_3$); Service Performance ($G_4$)

Three supply chain experts ($e_k, k = 1,2,3$) evaluate the alternative companies by using fuzzy natural language. In the following, we use the proposed intuitionistic fuzzy SIR method to solve this uncertainty group MCDM problem.

**Step 1.** Determine the individual measure degree $\xi_k$

The importance of supply chain experts is described using fuzzy natural language which is defined by intuitionistic fuzzy sets (see Table II). Table III provides the weights of expert assigned by Intuitionistic Fuzzy Numbers (IFNs). We use Eq (3) and Eq (4) to calculate the individual measure degree:

$\xi_k = (1.0000, 0.8314, 0.7405)$

**Step 2.** Group integration process

Table IV shows the experts' individual decision matrix, and Table V shows the weights assigned to every criterion, which are all described by fuzzy natural language terms.

| Alternative Partner | Expert | Criterion | | | |
|---|---|---|---|---|---|
| | | $G_1$ | $G_2$ | $G_3$ | $G_4$ |
| $Y_1$ | $e_1$ | VP | P | VVP | VP |
| | $e_2$ | VVP | VP | VP | VP |
| | $e_3$ | VP | P | VVP | VP |
| $Y_2$ | $e_1$ | P | M | VP | P |

| Expert | "Importance" Measure | IFNs $(\mu_k, v_k, \pi_k)$ |
|---|---|---|
| $e_1$ | Extremely Important | (1.0, 0.0, 0.0) |
| $e_2$ | Very Important | (0.8, 0.1, 0.1) |
| $e_3$ | Important | (0.7, 0.2, 0.1) |

Table III. The weight of experts on term "Importance"

|     |       |     |     |     |     |
|-----|-------|-----|-----|-----|-----|
|     | $e_2$ | VP  | P   | P   | M   |
|     | $e_3$ | M   | VP  | P   | P   |
|     | $e_1$ | AP  | VVP | VVP | VVP |
| $Y_3$ | $e_2$ | VVP | VP  | VVP | VVP |
|     | $e_3$ | VVP | VVP | VP  | VP  |
|     | $e_1$ | P   | M   | VVP | VP  |
| $Y_4$ | $e_2$ | VP  | M   | VP  | P   |
|     | $e_3$ | VP  | P   | VP  | P   |
|     | $e_1$ | M   | N   | VP  | M   |
| $Y_5$ | $e_2$ | P   | M   | VP  | P   |
|     | $e_3$ | P   | M   | P   | M   |

Table IV. Individual decision matrix on term "Quality"

| Expert | Weight of criterion | | | |
|---|---|---|---|---|
|  | $\omega_1$ | $\omega_2$ | $\omega_3$ | $\omega_4$ |
| $e_1$ | GI | I | VI | M |
| $e_2$ | VI | M | I | I |
| $e_3$ | I | I | M | M |

Table V. Criteria weights on term "Importance"

Based on Table II and Table IV, we calculate the group-integrated decision matrix with Eq (5).

$$\bar{d}_{ij} = (\bar{\mu}_{ij}, \bar{v}_{ij}) = \begin{bmatrix} (0.9677, 0.0090) & (0.9254, 0.0323) & (0.9777, 0.0048) & (0.9548, 0.0159) \\ (0.9120, 0.0399) & (0.9043, 0.0446) & (0.9289, 0.0301) & (0.8859, 0.0574) \\ (0.9920, 0.0027) & (0.9777, 0.0048) & (0.9785, 0.0045) & (0.9785, 0.0045) \\ (0.9397, 0.0239) & (0.8574, 0.0766) & (0.9699, 0.0080) & (0.9289, 0.0301) \\ (0.8816, 0.0603) & (0.7982, 0.1184) & (0.9441, 0.0215) & (0.8603, 0.0746) \end{bmatrix}$$

Based on Table II and Table V, we calculate the group-integrated weights of criteria with Eq (6).
$\bar{\omega}_j = (\bar{\mu}_j, \bar{v}_j) = ((0.9892, 0.0022), (0.9309, 0.0284), (0.9560, 0.0133), (0.8900, 0.0532))$

**Step 3.** Obtain intuitionistic fuzzy S-matrix and I-matrix
a) We use Eq (7) and Eq (8) to obtain the performance function $g_k(Y_i)$:

$$g_k(Y_i) = \psi(d_{ij}, d^+) = \begin{array}{c} \\ Y_1 \\ Y_2 \\ Y_3 \\ Y_4 \\ Y_5 \end{array} \begin{bmatrix} G_1 & G_2 & G_3 & G_4 \\ (0.9684) & (0.9284) & (0.9781) & (0.9561) \\ (0.9160) & (0.9090) & (0.9317) & (0.8920) \\ (0.9920) & (0.9781) & (0.9789) & (0.9789) \\ (0.9418) & (0.8662) & (0.9706) & (0.9317) \\ (0.8881) & (0.8137) & (0.9460) & (0.8688) \end{bmatrix}$$

b) We set the threshold criterion function as
$$\phi_k(d) = \begin{cases} 0.01 & \text{if } d > 0 \\ 0.00 & \text{if } d \leq 0 \end{cases}$$

**Step 4.** Obtain S-flow and I-flow
The S-flow can now be calculated using Eq (12) and the I-flow can be calculated using Eq (13) (see Table VI).

| Alternative | S-flow $(\phi^>(Y_i))$ | $s(\phi^>(Y_i))$ | I-flow $(\phi^<(Y_i))$ | $s(\phi^<(Y_i))$ |
|---|---|---|---|---|
| $Y_1$ | (0.3134, 0.6017) | -0.2883 | (0.1178, 0.8442) | -0.7264 |
| $Y_2$ | (0.1138, 0.8506) | -0.7368 | (0.3164, 0.5971) | -0.2807 |
| $Y_3$ | (0.3942, 0.5079) | -0.1137 | (0.0000, 1.0000) | -1 |
| $Y_4$ | (0.1636, 0.7852) | -0.6216 | (0.2758, 0.6469) | -0.3711 |
| $Y_5$ | (0.0308, 0.9577) | -0.9269 | (0.3750, 0.5304) | -0.1554 |

Table VI. S-flow and I-flow of alternatives

**Step 5.** The IF-SIR ranking
Based on the data in Table VI, we use Eq (2) and the IF-SIR rules to rank alternative partners.
We get the descending order of S-flow as: $s(\phi^>(Y_3)) > s(\phi^>(Y_1)) > s(\phi^>(Y_4)) > s(\phi^>(Y_2)) > s(\phi^>(Y_5))$
Therefore, the superiority ranking can be obtained as: $R_>^*: \{Y_3\} \to \{Y_1\} \to \{Y_4\} \to \{Y_2\} \to \{Y_5\}$
We get the ascending order of I-flow as: $s(\phi^<(Y_3)) < s(\phi^<(Y_1)) < s(\phi^<(Y_4)) < s(\phi^<(Y_2)) < s(\phi^<(Y_5))$

Therefore, the inferiority ranking can be obtained as: $R_<^*: \{Y_3\} \to \{Y_1\} \to \{Y_4\} \to \{Y_2\} \to \{Y_5\}$

**Step 6.** Map the complete ranking and make decision

Combined with the S-ranking and I-ranking into ranking structure $R^* = \{P, I, R\} = R_>^* \cap R_<^*$, we map the complete IF-SIR ranking from superior to inferior as: $\{Y_3\} \to \{Y_1\} \to \{Y_4\} \to \{Y_2\} \to \{Y_5\}$. Therefore, Partner $Y_3$ should thus be selected as the appropriate supply chain partner.

## 5. CONCLUSION

This study proposes a novel intuitionistic fuzzy SIR method to solve the uncertainty group multi-criterion decision making problem. We apply the intuitionistic fuzzy sets to define the fuzzy natural language terms which are used to describe the individual decision values and the weights for criteria and for decision makers. We first carry out an analysis of group MCDM problem to form a general model, and then expound our proposed method in solving uncertainty group MCDM problem. Finally, a practical application related to supply chain management is demonstrated. In future work, we will consider how to combine our proposed method with other uncertainty theories such as Rough sets and Grey measure to solve more uncertainty decision making problems in business or industry areas.

**Acknowledgement**

The authors would like to acknowledge the partial supports from the GRF 5237/08E of the Hong Kong Polytechnic University.

**RERFERENCES**